# Point Context: An Effective Shape Descriptor for RST-invariant Trajectory Recognition

Xingyu Wu[1], Xia Mao[1], Lijiang Chen[1], Yuli Xue[1] and Alberto Rovetta[2]

[1]School of Electronics and Information Engineering, Beihang University
Beijing, 100191, China (e-mail: moukyou@buaa.edu.cn)
[2]Department of Mechanics, Polytechnic University of Milan, Via La Masa 34, Milan 20156, Italy

*Abstract*—Motion trajectory recognition is important for characterizing the moving property of an object. The speed and accuracy of trajectory recognition rely on a compact and discriminative feature representation, and the situations of varying rotation, scaling and translation has to be specially considered. In this paper we propose a novel feature extraction method for trajectories. Firstly a trajectory is represented by a proposed *point context*, which is a rotation-scale-translation (RST) invariant shape descriptor with a flexible tradeoff between computational complexity and discrimination, yet we prove that it is a complete shape descriptor. Secondly, the *point context* is nonlinearly mapped to a subspace by Kernel Nonparametric Discriminant Analysis (KNDA) to get a compact feature representation, and thus a trajectory is projected to a single point in a low-dimensional feature space. Experimental results show that, the proposed trajectory feature shows encouraging improvement than state-of-art methods.

## 1 Introduction

Motion trajectory provides information about an object's movement patterns over time. The analysis and recognition of trajectory has become an attractive research field in recent years. A key problem in trajectory recognition is concerning the *rotation*, *scale* and *translation* (RST) invariant issue. In many applications such as visual surveillance [1], action recognition [2] and gesture interaction [3, 4], a moving pattern can be presented from various viewpoints and positions. These variations are modeled as affine transformation for 2D trajectories, and RST transformation for 3D trajectories in a Cartesian coordinate system. These transformations will cause mismatching when comparing a perceived trajectory with the memorized ones. Therefore, a trajectory recognition system needs to work in a manner that can eliminate the effect of viewpoint and position changes. Compared with 2D trajectories, richer information is contained in 3D trajectories and therefore better performance can be achieved [5]. So in this paper we focus on the RST-invariant 3D trajectory recognition problem.

A few researches have been conducted on this problem. One feasible approach is to represent a trajectory as the combination of RST-invariant basis functions. Reference [6] used Discrete Cosine Transform (DCT) to decompose a 3D trajectory into a set of coefficients, and the result is invariant to rotation and translation. Reference [7] added a multichannel sparse approximations process on the basis of [6] to estimate the optimized combination, and the performance was further enhanced. Other decomposition methods include the Fourier descriptor [8], the wavelet descriptor [9], Bezier curve [10] and invariant moments [11]. These methods cannot be applied to more generalized situations, as they have to tradeoff between the coarse (low frequency) and detailed (high frequency) information. Some other works used multi-view methods to find the correspondence between different viewpoints, in which a rotation-invariant fundamental matrix is estimated [12, 13]. These methods require a complex multi-camera setup.

Another kind of researches tries to extract the RST-invariant features of trajectories. These features can be categorized as the shape feature, as shape is inherently RST-invariant. *Curvature* and *torsion* are two widely used shape features for trajectory recognition [14, 15, 16], due to their sensitivity on local changes. Reference [17] used Curvature Scale Space (CSS) to associate the curvature zero-crossing points to the scale factor, and the coordinates of merging positions are then used for trajectory matching. Reference [18] compared CSS with Centroid Distance Function (CDF), and got their respective results. Reference [19] used a mixture of curvature, torsion and their first order derivatives as the descriptor. Reference [4] added an additional global item on the basis of [19] to create a mixed descriptor and slightly enhanced the performance. The main disadvantage of using *curvature* and *torsion* is that higher-order derivative is required in the calculation (two-order for *curvature* and three-order for *torsion*), so they are sensitive to outliers and fluctuations. Reference [20] proposed shape context, which utilized a polar histogram around each trajectory point, and each bin counts the number of other points in this area. Shape context shows its effectiveness in many works [21, 22, 23]. Reference [3] integrated shape context with motion direction, which is then decomposed by harmonic basis functions to be invariant to horizontal rotation. One main weakness of shape context is that, although it is a rich descriptor, abundant information is used to describe a single trajectory point without a deeper insight of which information is needed and which is not. Hence it brings unnecessary computation burden.

In this paper, we are the first to introduce the concept of complete shape descriptor, which means a descriptor that can preserve all the shape information of a trajectory and corresponds to a single kind of shape. Therefore, the complete shape descriptor has the best ability in distinguishing different trajectory shapes. Then we derive two kinds of complete shape descriptors with different lengths, based on which a *point context* descriptor is proposed. The *point context* has a compact form, yet we prove that it is a complete shape descriptor, with a flexible trade-off between the recognition accuracy and computational complexity.

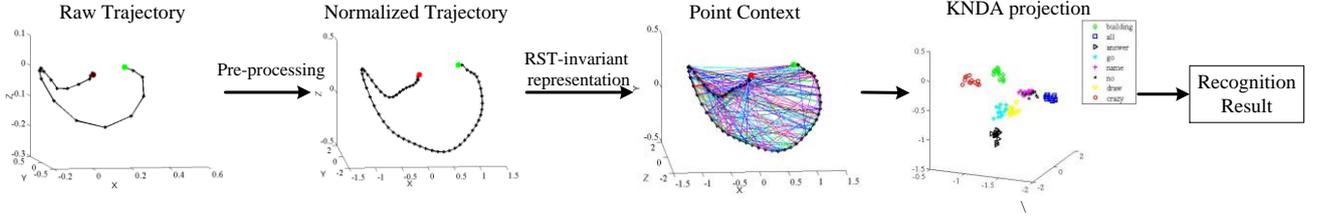

Fig. 1 Framework of the proposed method.

We concatenate the *point context* representation to a high-dimensional feature vector, which is then nonlinearly mapped to a low-dimensional subspace by Kernel Nonparametric Discriminant Analysis (KNDA) [24, 25]. Thus the trajectory can be recognized by a classifier on the subspace. Experimental results show the RST-invariance of our approach with the *point context* representation and KNDA mapping, which outperform state-of-the-art methods.

The remaining parts of this paper are organized as follows: Section 2 presents details of the proposed method, including pre-processing, the theory of *point context*, and nonlinear mapping. Section 3 presents experiments and result analysis. This paper is concluded in Section 4.

## 2 Trajectory Representation

The framework of the proposed approach is shown in Fig. 1. We expatiate each step in the following parts.

### 2.1 Pre-processing

Robust representation for trajectories must be invariant to the unique characteristics of 3D trajectory data, such as different moving speed or sampling rates, outliers, and various sequence lengths. These variations are eliminated in the pre-processing stage. Firstly, resampling along the trajectories is conducted to ensure that all trajectories consist of the same number of evenly distributed points. We define a 3D trajectory of $N$ sampling points as

$$\{\mathbf{p}'_l\}_{l=1}^N = \{x'_l, y'_l, z'_l\}_{l=1}^N, \tag{1}$$

where $x'_l$, $y'_l$ and $z'_l$ are the $X$, $Y$, and $Z$ coordinates of the $l$-th trajectory point $\mathbf{p}'_l$, respectively. These points are not evenly distributed due to the variations of moving speed or sampling rate, as shown in Fig. 2(a). Besides, the number of sampling points may differ across trajectories. So linear re-sampling is firstly employed to insert $n$ points evenly along the trajectory, and transfer the trajectory $\{\mathbf{p}'_l\}_{l=1}^N$ to $\{\mathbf{p}''_i\}_{i=1}^n$:

$$\begin{cases} \mathbf{p}''_1 = \mathbf{p}'_1 \\ \mathbf{p}''_{i+1} = \mathbf{p}''_i + \mathbf{\eta} \bullet L \end{cases}, \quad i = 1, 2, \cdots, n, \tag{2}$$

where $L$ is the average Euclidean distance between two consecutive points along the original trajectory, and $\mathbf{\eta}$ is the direction vector from $\mathbf{p}'_l$ to the next point of $\{\mathbf{p}''_i\}_{i=1}^n$

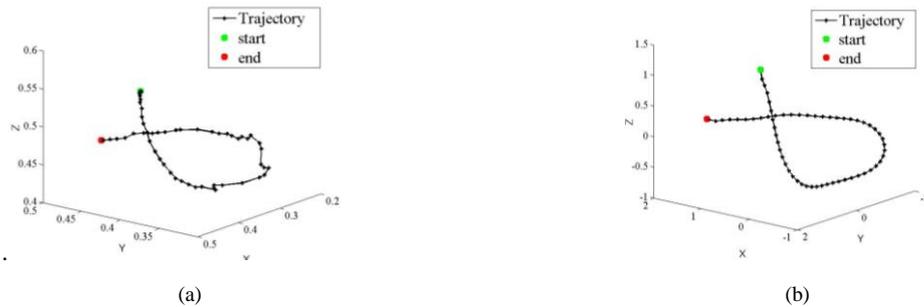

(a)            (b)

Fig. 2. An illustration of pre-processing. (a) The raw trajectory; (b) The trajectory after re-sampling, normalization and wavelet smoothing.

Secondly, the re-sampled trajectory $\{\mathbf{p}''_i\}_{i=1}^n$ is normalized to remove the effect of translation and scaling. The derived trajectory

is indicated by $\{\mathbf{p}_i\}_{i=1}^{n}$:

$$\mathbf{p}_i = \frac{\mathbf{p}_i'' - \frac{1}{n}\sum_{i=1}^{n}\mathbf{p}_i''}{\frac{1}{n}\sum_{i=1}^{n}\left\|\mathbf{p}_i'' - \frac{1}{n}\sum_{i=1}^{n}\mathbf{p}_i''\right\|}. \tag{3}$$

Finally, the trajectory $\{\mathbf{p}_i\}_{i=1}^{n}$ is smoothed by using a wavelet smoother. Trajectory smoothing is an important process to reduce the noise in a trajectory. The wavelet smoother introduced by [19] shows a desired balance between noise reduction and shape preservation, so it is adopted in our research. We used the same DB4 wavelet as in [19]. An example of the after-processing trajectory is shown in Fig. 2(b), which is the input of the next stage.

*2.2 Point Context Representation*

Now we need to get a RST-invariant representation of the trajectory. We introduce the shape descriptor to solve this problem. What's the essence of shape? We use the definition in [26]:

**Definition 1**. *Shape is all the geometrical information that remains when the translations, scale, reflect and rotational effects are filtered out from an object.*

By Definition 1, shape is feasible for RST-invariant representation. In this subsection we will give a detailed derivation for how much information is required to characterize the shape of a trajectory. Based on this theory, we then propose a new shape descriptor named *point context*.

**Lemma 1.** *Given a set of n different points $\{\mathbf{p}_i\}_{i=1}^{n}$ in a 3D Cartesian coordinate system. Let $d(\mathbf{p}_i, \mathbf{p}_j)$ denote the Euclidean distance between two points $\mathbf{p}_i$ and $\mathbf{p}_j$. If a mapping $f(\bullet)$ satisfies*

$$d(f(\mathbf{p}_i), f(\mathbf{p}_j)) = d(\mathbf{p}_i, \mathbf{p}_j),\ i=1,\ldots,n,\ j=1,\ldots,i, \tag{4}$$

*then $f(\bullet)$ is an isometric mapping, which does not change the shape of $\{\mathbf{p}_i\}_{i=1}^{n}$.*

Indeed, an isometric mapping in a 3D Cartesian coordinate system can be decomposed into a combination of rotation, translation and reflection transformations, so the shape is not changed. Therefore, we can get the following theorem:

**Theorem 1.** *Suppose that $\{\mathbf{p}_i\}_{i=1}^{n}$ has been normalized to remove the effect of scaling. Let*

$$C(n) = \{d(\mathbf{p}_i, \mathbf{p}_j)\}, i=1,\ldots,n,\ j=1,\ldots,i \tag{5}$$

*denote the Euclidean distances among all the n points, then the shape of $\{\mathbf{p}_i\}_{i=1}^{n}$ is determined by $C(n)$.*

**Proof.** Suppose there exists another trajectory $\{\hat{\mathbf{p}}_i\}_{i=1}^{n}$ whose $\hat{C}(n)$ is the same as $C(n)$. We assume a mapping $f:\{\mathbf{p}_i\}_{i=1}^{n} \mapsto \{\hat{\mathbf{p}}_i\}_{i=1}^{n}$ that satisfies

$$\begin{aligned}\hat{C}(n) &= \{d(\hat{\mathbf{p}}_i, \hat{\mathbf{p}}_j)\} \\ &= \{d(f(\mathbf{p}_i), f(\mathbf{p}_j))\}, i=1,\ldots,n,\ j=1,\ldots,i\end{aligned} \tag{6}$$

According to Lemma 1, $f(\bullet)$ must be an isometric mapping. So $\{\hat{\mathbf{p}}_i\}_{i=1}^{n}$ has the same shape with $\{\mathbf{p}_i\}_{i=1}^{n}$. Therefore, $C(n)$ determines the shape of $\{\mathbf{p}_i\}_{i=1}^{n}$, Theorem 1 is proved.

We call shape descriptors like $C(n)$ the complete shape descriptor, as they contain all the shape information of a point set. Different from the incomplete shape descriptors that correspond to more than one shape, the complete shape descriptors correspond to a single shape. Therefore, they have theoretically the best ability in distinguishing different trajectories. Figure 3 shows the limitations of some popular incomplete shape descriptors, including curvature, CSS and CDF, which cannot discriminate between certain kinds of trajectory shapes. However, these limitations do not occur on the complete shape descriptor.

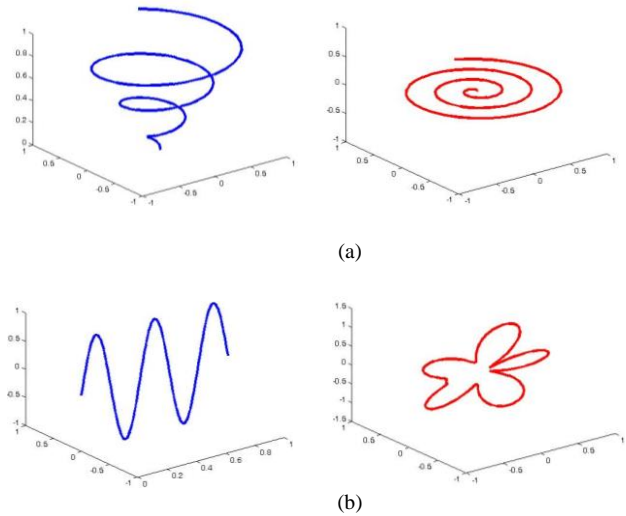

(a)

(b)

Fig 3. The limitations of some incomplete shape descriptors. (a) Cannot be distinguished by curvature and CSS; (b) Cannot be distinguished by CDF.

The dimension of $C(n)$ is $n(n-1)/2$ while the raw trajectory's dimension is $3n$ (three coordinates per point). This means that the space complexity of $C(n)$ is $O(n^2)$ while that of the raw data is $O(n)$. Therefore, much abundant information is contained in $C(n)$. In the following we will derive a more compact complete shape descriptor.

**Lemma 2.** *Given k spheres in a 3D space whose centers are non-coplanar, they can have only one point of intersection at most when $k \geq 4$.*

**Theorem 2.** *Suppose $\{\mathbf{p}_i\}_{i=1}^{n}$ has been normalized to remove the effect of scaling. For any $m \in [5..n]$, select from the set $\{\mathbf{p}_i\}_{i=1}^{m-1}$ four points $\mathbf{p}_{(1)}^m, \mathbf{p}_{(2)}^m, \mathbf{p}_{(3)}^m, \mathbf{p}_{(4)}^m$ that satisfy the following condition:*

*Condition 1: The points $\mathbf{p}_{(1)}^m,\ldots,\mathbf{p}_{(4)}^m$ have the same spatial relationship with $\{\mathbf{p}_i\}_{i=1}^{m-1}$, i.e., are both non-coplanar, coplanar or collinear.*

*We use $\mathbf{c}_m$ to represent the Euclidean distances from $\mathbf{p}_m$ to the four selected points*

$$\mathbf{c}_m = \{d(\mathbf{p}_m,\mathbf{p}_{(1)}^m), d(\mathbf{p}_m,\mathbf{p}_{(2)}^m), d(\mathbf{p}_m,\mathbf{p}_{(3)}^m),\ldots,d(\mathbf{p}_m,\mathbf{p}_{(4)}^m)\}, \qquad (7)$$

*and then*

$$C(n,4) = C(4) \cup \{\mathbf{c}_5,\ldots,\mathbf{c}_m,\ldots,\mathbf{c}_n\} \qquad (8)$$

*is a complete shape descriptor of $\{\mathbf{p}_i\}_{i=1}^{n}$, as illustrated in Fig. 4, where $C(4)$ is calculated by Eq. (5). The proof is given in Appendix A.*

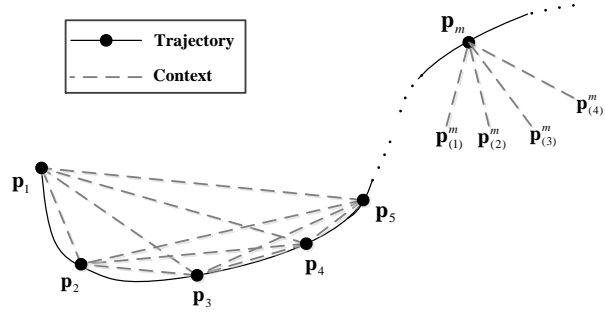

Fig 4. The shape descriptor $C(n, 4)$ is computed by the Euclidean distances from a certain point to four of its context points.

The space complexity of $C(n,4)$ is $O(n)$. Therefore, $C(n,4)$ is a compact complete shape descriptor. When $n \to \infty$, the dimension of $C(n,4)$ goes to $4n$ while that of the raw trajectory is $3n$. This means that $n$ additional dimensions are required to achieve the RST-invariance. In other words, the removing of the translation, scale, and rotation effect of a 3D trajectory needs an additional DOF (degree of freedom) for each point. This is a quantitative analysis of Definition 1.

Theorem 2 can be explained from another view. Suppose we compute the distance between two different points in $\{\mathbf{p}_i\}_{i=1}^n$, then we have $n(n-1)/2$ different choices, but only $4n$ among them are needed for a complete shape descriptor. Based on this theory, the dimensionality of complete shape descriptor can be compressed without information loss. As to the classification task, the discrimination can be increased by adding some abundant information. Therefore, we define a parameter $\lambda \in [1..n]$ to change Eq. (8) as

$$C(n,\lambda) = \begin{cases} C(\lambda) \cup \{\mathbf{c}_{\lambda+1},...,\mathbf{c}_n\} & if \quad \lambda < n \\ C(n) & if \quad \lambda = n \end{cases}. \qquad (9)$$

when $\lambda > 4$, it is obvious that $C(n,4) \subseteq C(n,\lambda)$, so $C(n,\lambda)$ is a complete shape descriptor. Note that $C(n,\lambda)$ is equal to $C(n)$ when $\lambda = n$, so Theorem 1 can be regarded as a special case of Theorem 2. Indeed, smaller $\lambda$ can reduce the computational complexity, and larger $\lambda$ will increase the stability with additional information. As each point is represented by their $\lambda$ context points, we name the shape descriptor $C(n,\lambda)$ as *point context*.

Another important concept is the *context table*. To keep the correspondence of dimensions, a look-up table should be shared by all trajectories in the sample set to determine the indexes of the $\lambda$ contexts for each point, i.e. to determine which $\lambda$ points are used to compute their distances from a certain point. We define a $n*\lambda$ matrix $\mathbf{M}$, whose element $m_{ij} = a$ indicates that $\mathbf{p}_a$ is one of the $j$ context point of $\mathbf{p}_i$. Thus $\mathbf{M}$ can be used to determine how to select the context points. The element $m_{ij}$ of $\mathbf{M}$ needs to satisfy the condition:

$$m_{ij} \in [1..i-1] \qquad (10)$$

It is worth noting that here we loosen the restriction of *Condition 1*, which cannot be guaranteed when all samples share the same *context table*. Nevertheless, in most cases *Condition 1* can be satisfied as the $\lambda$ context points are randomly selected, especially when $\lambda$ is large. Some special cases will not downgrade the performance a lot in practice. In our previous work [27], the proposed method is equal to set $\lambda$ as $n$, which proves to be a redundant representation that brings heavy computation burden while contributing little to the precision.

2.3  *Kernel Representation*

After the above processing, each trajectory is represented by a feature vector $\mathbf{v} \in R^h$, whose dimension $h$ is the length of $C(n,\lambda)$. The vector has to be projected to a low-dimensional feature space for feature extraction. Meanwhile, suppose a training set $\{\mathbf{v}_r\}_{r=1}^R$, in many situations the sample dimension $h$ exceeds the sample number $R$, which makes the dimensionality reduction a small sample size problem [28]. In this subsection we employ a kernel-based method to solve the dimensionality reduction problem.

Kernel methods have been successfully applied in many fields. They work by employing a nonlinear mapping $\phi(\cdot)$ from the original input space $R^h$ to an arbitrarily large or infinite dimensional kernel space $F$. The linear operations in $F$ are dual to the nonlinear operations in $R^h$. We use KNDA in this research as the dimensionality reduction algorithm. In this subsection, we first review the LDA framework, and then KNDA is introduced and compared.

The traditional LDA algorithm finds the most discriminant subspace by maximizing the between-class scatter while minimizing the within-class scatter. Suppose there are $C$ pattern classes, let $\{\mu_s\}_{s=1}^{C}$ be the center of the $s$-th class, and $\{N_s\}_{s=1}^{C}$ be the number of samples in the $s$-th class. The within-class scatter matrix of training set $\mathbf{V} = \{\mathbf{v}_r\}_{r=1}^{R}$ is defined as

$$S_w = \frac{1}{N} \sum_{s=1}^{C} \sum_{r=1}^{N_s} (\mathbf{v}_r - \mathbf{\mu}_s)(\mathbf{v}_r - \mathbf{\mu}_s)^T, \tag{11}$$

and the between-class scatter matrix is defined as

$$S_b = \frac{1}{R} \sum_{s=1}^{C} (\mathbf{\mu}_s - \mu)(\mathbf{\mu}_s - \mu)^T, \tag{12}$$

where $\mu$ is the center of the entire training set. We assume a linear mapping $\mathbf{w}: R^h \to R^l$ from the sample space to a low-dimensional subspace, and the projected scatter matrices $\tilde{S}_w$ and $\tilde{S}_b$ in $R^l$ are calculated by

$$\tilde{S}_w = \mathbf{w}^T S_w \mathbf{w} \tag{13}$$

and

$$\tilde{S}_b = \mathbf{w}^T S_b \mathbf{w}. \tag{14}$$

The optimal linear projection $\mathbf{w}$ can be found by maximizing the Fisher Linear Discriminant

$$J(\mathbf{w}) = \frac{|\tilde{S}_b|}{|\tilde{S}_w|} = \frac{\mathbf{w}^T S_b \mathbf{w}}{\mathbf{w}^T S_w \mathbf{w}}, \quad \mathbf{w} \neq 0. \tag{15}$$

Mathematically, $\mathbf{w}$ is equivalent to the $l$ leading eigenvectors of $S_w^{-1} S_b$.

However, the LDA algorithm has two major limitations. On one hand, it suffers performance degradation in cases of non-Gaussian distribution. On the other hand, the rank of $S_b$ in Eq. (12) is $C$-1, so the dimension of the constructed subspace $R^l$ is at most $C$-1, which is usually insufficient to separate different classes.

Nonparametric Discriminant Analysis (NDA) is proposed to alleviate the limitations of LDA [29], which redefines Eq. (12) as a nonparametric form that utilizes the whole dataset rather than merely the sample centers:

$$\hat{S}_b = \frac{1}{N} \sum_{s=1}^{C} \sum_{\substack{t=1 \\ t \neq s}}^{C} \sum_{r=1}^{N_s} w(s,t,r)(\mathbf{v}_r^s - \mathbf{m}_t(\mathbf{v}_r^s))(\mathbf{v}_r^s - \mathbf{m}_t(\mathbf{v}_r^s))^T, \tag{16}$$

where $\mathbf{v}_r^s$ denotes the $r$-th sample of the class $s$, $w(s,t,r)$ is a weight function that emphasizes samples near the classification boundary, and $\mathbf{m}_t(\mathbf{v}_r^s)$ is the local $k$-NN mean of $\mathbf{v}_r^s$ in class $t$, as shown in Fig. 5. The rank of $\hat{S}_b$ is equal to the sample number, thus the maximum dimension of $R^l$ is raised. Besides, as the entire sample information is utilized rather than just the centers, the NDA algorithm is more suited to non-Gaussian distribution than LDA.

However, NDA is still a linear dimensionality reduction algorithm that cannot reflect the nonlinear relationship between different dimensions. Therefore, it can be extended by kernel method to get a nonlinear performance. Suppose a nonlinear mapping $\phi(\cdot)$ from the input space to an arbitrarily large or infinite dimensional kernel space

$$\begin{aligned}\phi: R^h &\to F \\ \mathbf{v} &\mapsto \phi(\mathbf{v})\end{aligned}, \tag{17}$$

then the dot product of two vectors in the kernel space can be computed from the input space by the kernel function:

$$k(\mathbf{v}_1, \mathbf{v}_2) = \phi(\mathbf{v}_1)^T \phi(\mathbf{v}_2). \tag{18}$$

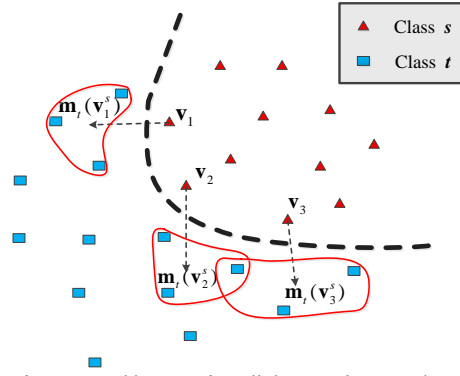

Fig. 5 The nonparametric between-class scatter is computed by counting all the samples over the whole dataset (here the 3 nearest neighbors are used).

A kernel function needs to satisfy the Mercer's condition [30]. Consider the Fisher discriminant in Eq. (15), according to the theory of reproducing kernels [31], the mapping $\mathbf{w}^\phi : F \to R^l$ can be expressed by the linear combination of the mapped samples

$$\mathbf{w}^\phi = \sum_{r=1}^{R} \alpha_r \phi(\mathbf{v}_r) = \boldsymbol{\alpha}\phi(\mathbf{V})^T \quad (19)$$

where $\boldsymbol{\alpha} = [\alpha_1, \ldots, \alpha_R]^T$ are the expansion coefficients. Therefore, Eq. (15) can be redefined in terms of $\boldsymbol{\alpha}$

$$J(\mathbf{w}^\phi) = \frac{\mathbf{w}^{\phi T} \hat{S}_b^\phi \mathbf{w}^\phi}{\mathbf{w}^{\phi T} S_w^\phi \mathbf{w}^\phi} = \frac{\boldsymbol{\alpha}^T \mathbf{B} \boldsymbol{\alpha}}{\boldsymbol{\alpha}^T \mathbf{A} \boldsymbol{\alpha}} \quad (20)$$

where $\hat{S}_b^\phi$ and $S_w^\phi$ indicate the between-class and within-class scatter matrixes of NDA in the kernel space $F$, while

$$\mathbf{A} = \phi(\mathbf{V})^T S_w^\phi \phi(\mathbf{V}) \quad (21)$$

and

$$\mathbf{B} = \phi(\mathbf{V})^T \hat{S}_b^\phi \phi(\mathbf{V}) . \quad (22)$$

We define a $R*R$ Gram matrix $\mathbf{K} = k(\mathbf{V}, \mathbf{V})$, whose element is computed from the training samples

$$K_{pq} = \phi(\mathbf{v}_p)^T \phi(\mathbf{v}_q) = k(\mathbf{v}_p, \mathbf{v}_q), \quad (23)$$

then $\mathbf{A}$ and $\mathbf{B}$ can be expressed by $\mathbf{K}$ using the kernel trick, without actually performing the mapping $\phi(\cdot)$:

$$\mathbf{A} = \sum_{s=1}^{C} \mathbf{K}_s (\mathbf{I} - \mathbf{1}_{\frac{1}{N_s}}) \mathbf{K}_s^T \quad (24)$$

and

$$\mathbf{B} = \frac{1}{N} \sum_{s=1}^{C} \sum_{\substack{t=1 \\ t \neq s}}^{C} \sum_{r=1}^{N_s} w^\phi(s,t,r)(K_{sr} - \mathbf{NN}_{sr}^{t\,\phi} \mathbf{1}_{\frac{1}{k}})(K_{sr} - \mathbf{NN}_{sr}^{t\,\phi} \mathbf{1}_{\frac{1}{k}})^T \quad (25)$$

where $\mathbf{K}_s$ indicates the $s$-th column of $\mathbf{K}$, $\mathbf{NN}_{sr}^{t\,\phi}$ preserves the $k$-NN mean of $\mathbf{v}_r^s$ in class $t$, in the kernel space. The detailed derivations can be found in [25].

Substituting Eq. (24) and Eq. (25) into Eq. (20), we can get the expansion coefficients $\boldsymbol{\alpha}$, which is equivalent to the $l$ leading eigenvectors of $\mathbf{A}^{-1}\mathbf{B}$. To overcome the singularity problem that caused by the small sample effect when computing $\mathbf{A}^{-1}$, PCA is applied beforehand on the Gram Matrix $\mathbf{K}$ to compress its dimensions. The detailed algorithm of using KNDA is given as follows:
1. Compute the Gram matrix $\mathbf{K}$ by the training set ;
2. Project $\mathbf{K}$ to its PCA subspace. Adjust the PCA dimension to better reduce noise and get the PCA projection matrix $\mathbf{P}$;

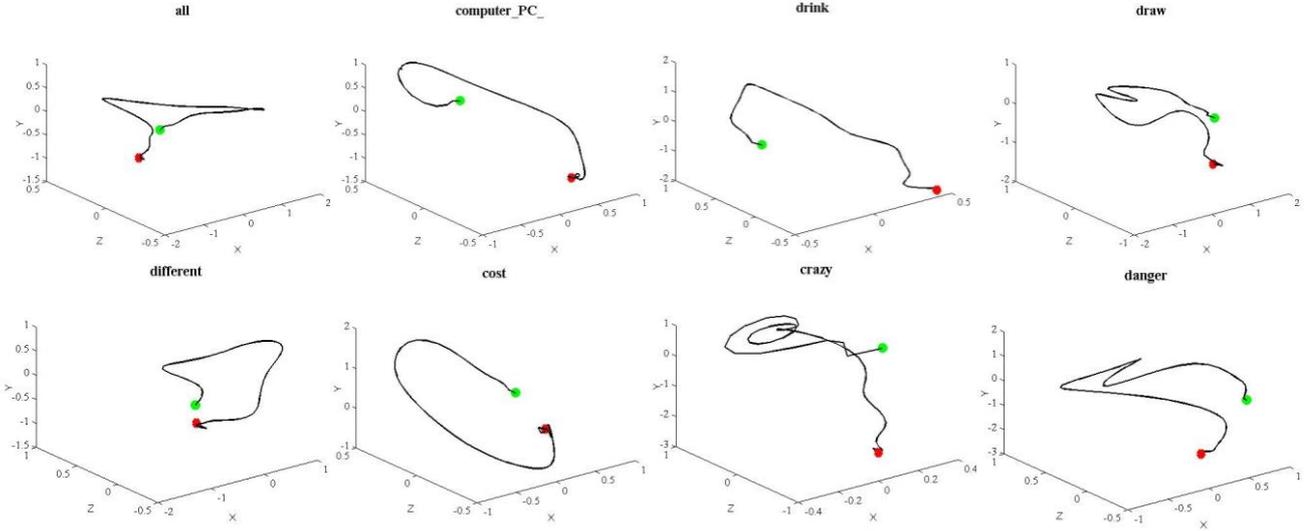

Fig. 6 The 3D trajectories of eight sign words from the ASL dataset.

3. Compute the matrix **A** and **B** as well as the expansion coefficients **α**.
4. Given a sample $\mathbf{u} \in R^h$, its corresponding feature vector $\mathbf{u}' \in R^l$ is computed by

$$\mathbf{u}' = \boldsymbol{\alpha}\phi(\mathbf{V})^T \phi(\mathbf{u})\mathbf{P} = \boldsymbol{\alpha} k(\mathbf{V},\mathbf{u})\mathbf{P} \tag{26}$$

The constructed feature subspace by KNDA is then classified by the Support Vector Machine (SVM) classifier [32] to get the recognition result.

## 3 Experimental Results

The evaluation of the proposed method mainly focuses on two parts. Firstly we evaluate the impact of different parameters on the *point context*. Secondly the RST-invariance is validated, and our results are compared with state-of-the-art methods.

We used the Australian Sign Language (ASL) [33] data set for experiment. It contains trajectory samples of 95 ASL signs performed by five different person. Each ASL sign is composed of 70 samples that record the position (consisting of the *X*, *Y* and *Z* coordinates), palm orientation and finger bend, and we only use the position information.

The involved parameters in our algorithm are set as follows:
The normalized trajectory length *n* in Eq. (2) is set as 60; The count of *k*-NN in Eq. (16) is chosen as the median of the sample number for each training class. The subspace dimension is set as *C* where *C* is the class number. For the KNDA algorithm and SVM classifier, the *RBF* kernel is chosen with the kernel parameter optimized by 10-fold cross-validation [34], in which the training set is divided into ten subsets equally (or nearly equally), then ten iterations are performed to test how the classification accuracy varies when different kernel parameters are used. In each iteration a different subset is left for validation while the remaining nine subsets are combined and used for training with different parameters. The parameter yielding the highest accuracy is kept. The experiment setup is PC (Intel Core i7 3370k, 4G RAM).

### 3.1 *The impact of context number and context table*

The proposed *point context* is actually a parametric method. In this part we evaluate the impact of different parameters. The data set consists of eight sign words 'all', 'computer', 'drink', 'raw', 'different', 'cost', 'crazy' and 'danger' from the ASL set, as shown in Fig. 6. Each sign-word category has 70 trajectories. Half trajectories from each category serve as training samples and the remaining serve as trials.

The context number $\lambda$ is traversed from 1 to 59. For each value of $\lambda$, the experiment is repeated 100 times with different context tables that are generated randomly under the constraint of Eq. (10). The average recognition rates obtained at different context numbers are shown in Fig. 7(a), where the error bar indicates one standard deviation. The time consumptions of training and testing process are shown in Fig. 7(b).

Figure. 7(a) illustrates the influence of $\lambda$ and **M** on the recognition rate, from which we can see that when $\lambda$ is small it is positively correlated to the recognition rate, and the variations of **M** cause more fluctuations. When $\lambda$ is large (typically bigger than *n/2*), its increasing makes no contribution to the recognition rate, and the variations of **M** also has little impact. Figure. 7(b) shows that $\lambda$ is always positive correlated to the time consumption of training and testing, because it is in proportion with the feature dimensions. Due to the above analysis, we can tune the parameter $\lambda$ to get a flexible trade-off between the accuracy and complexity, and use a random **M**.

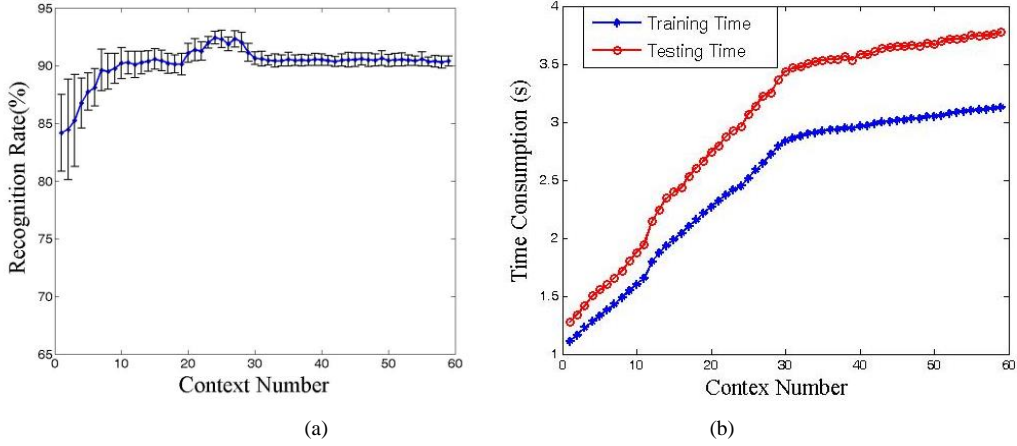

(a)                (b)

Fig. 7 The recognition results and time consumptions when using different parameters.

### 3.2 RST-invariance validation on a large date set

In this part we test the RST-invariant property of the proposed method, using the ASL data set. This experiment is composed of two parts. In the first part we use half samples in the data set for training, and the remaining half for testing. The test sample $\{x_i, y_i, z_i\}_{i=1}^n$ is transformed to $\{\hat{x}_i, \hat{y}_i, \hat{z}_i\}_{i=1}^n$ by a random RST transformation

$$\begin{bmatrix} \hat{x}_i & \hat{y}_i & \hat{z}_i \end{bmatrix}^T = \mathbf{t} + s\mathbf{R}\begin{bmatrix} x_i & y_i & z_i \end{bmatrix}^T \tag{28}$$

or

$$\begin{bmatrix} \hat{x}_i \\ \hat{y}_i \\ \hat{z}_i \end{bmatrix} = \begin{bmatrix} t_x \\ t_y \\ t_z \end{bmatrix} + s \begin{bmatrix} r_{11} & r_{12} & r_{13} \\ r_{21} & r_{22} & r_{23} \\ r_{31} & r_{32} & r_{33} \end{bmatrix} \begin{bmatrix} x_i \\ y_i \\ z_i \end{bmatrix}, \tag{29}$$

where $r_{ij}$ is the element of the orthogonal rotation matrix **R**, $\mathbf{t} = [t_x\ t_y\ t_z]^T$ indicates translation, and *s* is the scale factor. The parameters of **R**, *s* and **t** are randomly selected. The experiment is used to validate whether the test samples can be correctly recognized after a random RST transformation while the training samples are not transformed. The results are shown in TABLE I. In the second part, we remove the RST transformation process, and thus the testing samples are in the same RST state with the training samples, and the other setup stays the same. The results are shown in TABLE II.

Our results are recorded in two versions with $k=25$ and $k=59$, respectively, and compared with those reported in [8], [18], [19], [4] and [5]. We demonstrate the results with different classifiers including SVM, naive Bayes and *k*-NN. The entire ASL date set is used with 95 different categories of sign words, and we evaluate the cases of using two, four, eight, and sixteen classes, respectively. Each case is repeated 200 times with different sign-words constitution, to remove the effect of different selections, and the average recognition rate is recorded.

From TABLE I and TABLE II we can draw the following conclusions:
1. No significant change in recognition rate of the proposed method can be observed when the trials are changed by a RST transformation, compared with the non-transformed situation. So the proposed approach supports RST-invariant trajectory recognition. The deviations observed are mainly due to sample variations.
2. Different values of the context number do little change on the final recognition rate. This is in accordance with the result in Fig. 7(a).
3. The method proposed by [5] is actually a RST-dependent approach. TABLE I shows that the RST transformation pulls

down the recognition rate a lot, when compared with that of TABLE II. So RST-dependent approaches are not preferable when there exist RST changes in testing samples.

4. The proposed method outperforms the other RST-invariant methods, but is not better than [5] in Table I. This is because that the RST-invariance is achieved at the price of decreased accuracy. When different classes of trajectories have similar shapes and very different RST states, they can hardly be correctly classified by RST-invariant methods, as shown in Fig. 8. However, these classes can be easily distinguished by RST-dependent methods when the trials are not RST-transformed from the training models. So the average recognition rate of [5] is raised.

TABLE I Average recognition rate under RST transformation (%)

|  | Classifier | Invariance | Number of classes | | | |
|---|---|---|---|---|---|---|
|  |  |  | 2 | 4 | 8 | 16 |
| Harding *et al* [8] | PNN | RST-invariant | 87.95 | 75.74 | 63.85 | N/A |
| Bashir *et al* [18] | HMM | RST-invariant | 88.00 | 73.00 | N/A | N/A |
| Wu *et al* [19] | DTW | RST-invariant | 92.03 | 87.52 | 81.19 | N/A |
| Yang *et al* [4] | DTW | RST-invariant | 94.30 | 90.15 | 86.71 | N/A |
| Lin *et al* [5] | *k*-NN | RST-dependent | 81.32 | 58.72 | 45.68 | 38.71 |
| Proposed (k=25) | SVM | RST-invariant | 96.49 | **93.52** | **89.51** | **77.83** |
| Proposed (k=59) | SVM | RST-invariant | 96.02 | 93.11 | 88.30 | 77.56 |
| Proposed (k=25) | Bayes | RST-invariant | **96.82** | 92.77 | 88.42 | 76.85 |
| Proposed (k=59) | Bayes | RST-invariant | 95.07 | 92.65 | 88.20 | 76.45 |
| Proposed (k=30) | *k*-NN | RST-invariant | 95.45 | 92.70 | 86.89 | 74.31 |
| Proposed (k=59) | *k*-NN | RST-invariant | 94.83 | 91.58 | 85.71 | 74.48 |

TABLE II Average recognition rate without RST transformation (%)

|  | Classifier | Invariance | Number of classes | | | |
|---|---|---|---|---|---|---|
|  |  |  | 2 | 4 | 8 | 16 |
| Harding *et al* [8] | PNN | RST-invariant | 87.95 | 75.74 | 63.85 | N/A |
| Bashir *et al* [18] | HMM | RST-invariant | 88.00 | 73.00 | N/A | N/A |
| Wu *et al* [19] | DTW | RST-invariant | 92.03 | 87.52 | 81.19 | N/A |
| Yang *et al* [4] | DTW | RST-invariant | 94.30 | 90.15 | 86.71 | N/A |
| Lin *et al* [5] | *k*-NN | RST-dependent | 99.00 | 96.00 | 92.00 | 89.00 |
| Proposed (k=25) | SVM | RST-invariant | **96.71** | 93.87 | 89.77 | 79.58 |
| Proposed (k=59) | SVM | RST-invariant | 96.22 | 93.16 | 88.50 | 77.76 |
| Proposed (k=25) | Bayes | RST-invariant | 96.51 | 92.79 | 88.62 | 79.05 |
| Proposed (k=59) | Bayes | RST-invariant | 95.88 | 92.95 | 87.30 | 76.55 |
| Proposed (k=25) | k-NN | RST-invariant | 95.61 | 91.83 | 86.49 | 75.41 |
| Proposed (k=59) | k-NN | RST-invariant | 94.62 | 91.98 | 85.04 | 73.79 |

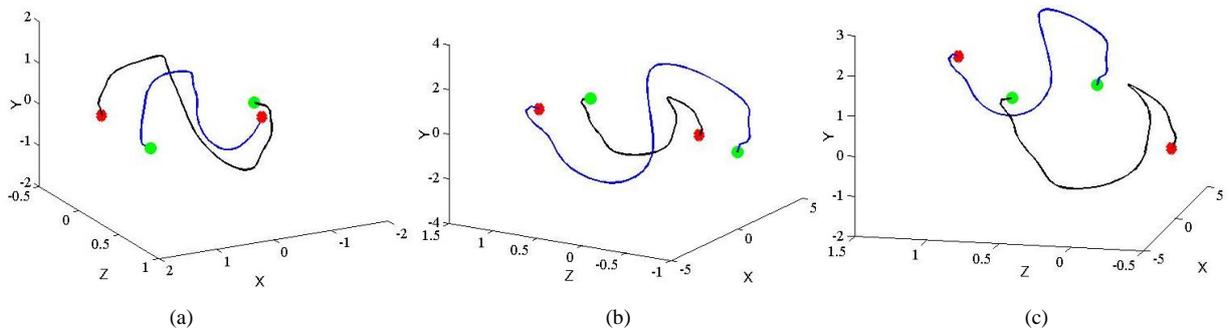

(a)      (b)      (c)

Fig. 8 The confusing trajectories that cannot be distinguished by RST-invariant methods. (a) Two trajectories differentiated by rotation. (b) two trajectories differentiated by scale. (c) two trajectories differentiated by translation.

# 4. Conclusion and Future Work

In this paper, we focus on the RST-invariant 3D trajectory recognition problem. The contributions of this paper are mainly twofold. Firstly, we prove that the shape of a trajectory can be derived by the distances between trajectory points. Based on this theory, we propose a parametric trajectory shape descriptor called *point context*, and the influence of different parameters is further analyzed. Secondly, we use a kernel-based discrimination analysis method to map the shape descriptor nonlinearly to a low-dimensional feature space. Consequently, we can extract the most compact yet discriminative features of a trajectory for classification. We validate the RST-invariant property of the proposed method on the ASL date set, and the results are compared with state-of-the-art approaches [8], [18], [19], [4].

A trajectory contains two kinds of information: the spatial information and the temporal information. In the future work, we may regard a trajectory as a time series rather than a feature vector, and the recognition can be achieved by series matching algorithms, such as Continuous Dynamic Time Warping [35] and Canonical Time Warping [36]. Series-based method supports online recognition, i.e., a trajectory can be recognized before it ends, by using only a portion of points. This will be important for many real-time applications.

## Acknowledgements

This work is supported by the National Natural Science Foundation of China (61103097), the Project supported by the Specialized Research Fund for the Doctoral Program of Higher Education of China (20121102130001) and the China Aerospace Support Technology Foundation.

# APPENDIX
# THE PROOF OF LEMMA 1
**Proof.**

1. Let $T_1 = \{\mathbf{p}_1\}$ be a trajectory that just contains a single point $\mathbf{p}_1$, then its shape is determined.

2. Let $T_2 = \{\mathbf{p}_1, \mathbf{p}_2\}$ where $\mathbf{p}_2 \neq \mathbf{p}_1$. Suppose

$$\exists \mathbf{p}_2' \quad s.t. \quad d_{12}' = d_{12}, \tag{30}$$

then $T_2' = \{\mathbf{p}_1, \mathbf{p}_2'\}$ can be transformed to $T_2$ by $\mathbf{Rot}(\mathbf{p}_1)$, which denotes a rotation around the point $\mathbf{p}_1$. So the shape of $T_2$ is determined by $\mathbf{p}_1$ and $d_{12}$, written as

$$\mathbf{p}_1 \cup d_{12} \to T_2. \tag{31}$$

3. Add another point $\mathbf{p}_3$ that satisfies $\mathbf{p}_3 \neq \mathbf{p}_1$ and $\mathbf{p}_3 \neq \mathbf{p}_2$. We use $sph(\mathbf{p}_i, d_{ij})$ to denote a sphere whose center is $\mathbf{p}_i$ and radius is $d_{ij}$. Suppose

$$\exists \mathbf{p}_3' \quad s.t. \quad d_{13}' = d_{13} \,\&\, d_{23}' = d_{23}, \tag{32}$$

then

$$\mathbf{p}_3, \mathbf{p}_3' \in sph(\mathbf{p}_1, d_{13}) \cap sph(\mathbf{p}_2, d_{23}). \tag{33}$$

The trajectory $T_3' = \{T_2, \mathbf{p}_3'\}$ can be transformed to $T_3$ by $\mathbf{Rot}(\mathbf{p}_1\mathbf{p}_2)$, which denotes a rotation around the axis $\mathbf{p}_1\mathbf{p}_2$, as illustrated in Fig. 9. Hence $T_3'$ and $T_3$ have the same shape. So the shape of $T_3$ is solely determined by $T_2$, $d_{13}$ and $d_{23}$, written as

$$T_2 \cup d_{13} \cup d_{23} \to T_3 \tag{34}$$

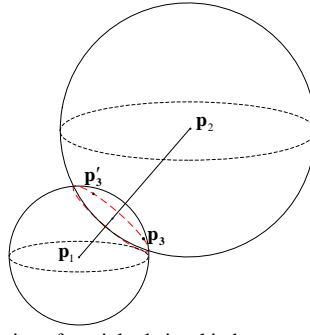

Fig. 9 An illustration of spatial relationship between three points.

4. Again add $\mathbf{p}_4$ that satisfies $\mathbf{p}_4 \neq \mathbf{p}_i$, $i=1,2,3$. then $T_4 = \{T_3, \mathbf{p}_4\}$. Suppose

$$\exists \mathbf{p}'_4 \quad s.t. \quad d'_{i4} = d_{i4}, \quad i=1,2,3, \tag{35}$$

then

$$\mathbf{p}_4, \mathbf{p}'_4 \in \bigcap_{i=1}^{3} Sph(\mathbf{p}_i, d_{i4}). \tag{36}$$

Now there exist the following two cases:

Case 1:

If $\mathbf{p}_1$, $\mathbf{p}_2$ and $\mathbf{p}_3$ are collinear, the situation is the same as in step 3, and $T'_4$ can be transformed to $T_4$ by $\mathbf{Rot}(\mathbf{p}_1\mathbf{p}_3)$, as shown in Fig. 10.

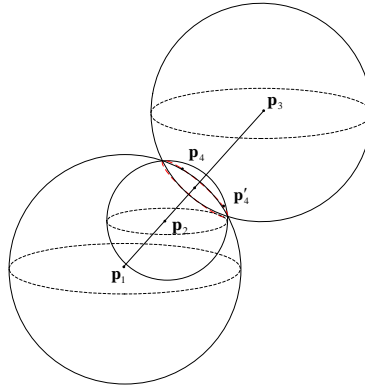

Fig. 10 An illustration of the spatial relationship between four points, when three of them are collinear.

Case 2:

If $\mathbf{p}_1$, $\mathbf{p}_2$ and $\mathbf{p}_3$ are non-collinear, $\mathbf{p}'_4$ is the specular point of $\mathbf{p}_4$, so $T'_4$ can be transformed to $T_4$ by $\mathbf{M}(p_1 p_2 p_3)$, which denotes a reflect transformation with respect to the plane consisting of the points $\mathbf{p}_1$, $\mathbf{p}_2$ and $\mathbf{p}_3$.

To summarize, $T'_4$ has the same shape with $T_4$, so the shape of $T_4$ is determined by $T_3$ and $\{d_{i4}\}_{i=1}^{3}$, written as

$$T_3 \cup d_{14} \cup d_{24} \cup d_{34} \to T_4. \tag{37}$$

5. Suppose that the shape of $T_m$ ($m \geq 4$) is determined by $T_{m-1}$ and $\{d_{im}\}_{i=1}^{m-1}$. We add another point $\mathbf{p}_{m+1}$ that satisfies

$$\mathbf{p}_{m+1} \neq \mathbf{p}_i, \quad i=1,2,\ldots,m, \tag{38}$$

then $T_{m+1} = \{T_m, \mathbf{p}_{m+1}\}$. Suppose

$$\exists \mathbf{p}'_{m+1} \quad s.t. \quad \mathbf{c}'_{m+1} = \mathbf{c}_{m+1}, \tag{39}$$

and then

$$\mathbf{p}_{m+1}, \mathbf{p}'_{m+1} \in \bigcap_{i=1}^{4} sph(\mathbf{p}_{(i)}^{m+1}, d(\mathbf{p}_{m+1}, \mathbf{p}_{(i)}^{m+1})). \tag{40}$$

Now there exist the following three cases:

Case 1:

If $\{\mathbf{p}_i\}_{i=1}^{m}$ are collinear, according to Condition 1, $\mathbf{p}_{(1)}^{m+1}, \mathbf{p}_{(2)}^{m+1}, \mathbf{p}_{(3)}^{m+1}, \mathbf{p}_{(4)}^{m+1}$ are collinear, the situation is the same as in step 3, and $T'_{m+1} = \{T_m, \mathbf{p}'_{m+1}\}$ can be transformed to $T_{m+1}$ by $\mathbf{Rot}(\mathbf{p}_{(1)}^{m+1}\mathbf{p}_{(4)}^{m+1})$;

Case 2:

If $\{\mathbf{p}_i\}_{i=1}^{m}$ are non-collinear but coplanar, the situation is the same as in step 4, and $T'_{m+1}$ can be transformed to $T_{m+1}$ by $\mathbf{M}(\mathbf{p}_{(1)}^{m+1}\mathbf{p}_{(2)}^{m+1}\mathbf{p}_{(3)}^{m+1})$;

Case 3:

If $\{\mathbf{p}_i\}_{i=1}^{m}$ are non-coplanar, according to Lemma 2, $\mathbf{p}'_{m+1} = \mathbf{p}_{m+1}$, so $T'_{m+1} = T_{m+1}$.

To summarize, $T'_{m+1}$ has the same shape with $T_{m+1}$, so the shape of $T_{m+1}$ is solely determined by $T_m$ and $\mathbf{c}_{m+1}$, written as

$$T_m \cup \mathbf{c}_{m+1} \to T_{m+1}. \tag{41}$$

6. As proved in step 1~step 5, we have

$$\begin{aligned} T_{n-1} \cup \mathbf{c}_n &\to T_n \\ T_{n-2} \cup \mathbf{c}_{n-1} \cup \mathbf{c}_n &\to T_n \\ &\vdots \\ \mathbf{p}_1 \cup C(4) \cup \mathbf{c}_5 \cup \ldots \cup \mathbf{c}_n &\to T_n \end{aligned} \tag{42}$$

So the shape of $T_n$ is solely determined by $\mathbf{p}_1 \cup C(4) \cup \{\mathbf{c}_5, \ldots, \mathbf{c}_m, \ldots, \mathbf{c}_n\}$.

7. Now consider the effect of $\mathbf{p}_1$. Suppose $\exists \mathbf{p}'_1$ s.t. $d'_{1i} = d_{1i}$, $i = 1, 2, \ldots, n$, the trajectory $T'_n$ can be transformed to $T_n$ by translation, so they share the same shape. Hence the change of $\mathbf{p}_1$ will not change the shape $T_n$. So The shape of $T_n$ is determined by

$$C(n, 4) = C(4) \cup \{\mathbf{c}_5, \ldots, \mathbf{c}_m, \ldots, \mathbf{c}_n\}. \tag{43}$$

Theorem 2 is proved.